%% file: main.tex
\documentclass[letterpaper, 10 pt, conference]{ieeeconf}  % Comment this line out if you need a4paper

\IEEEoverridecommandlockouts                              % This command is only needed if 
\usepackage{graphics} % for pdf, bitmapped graphics files
\usepackage{epsfig} % for postscript graphics files
\usepackage{hyperref}
\usepackage{balance}
\usepackage{mathptmx} % assumes new font selection scheme installed
\usepackage{times} % assumes new font selection scheme installed
\usepackage{amsmath} % assumes amsmath package installed
\usepackage{amssymb}  % assumes amsmath package installed
\usepackage{interval}
\usepackage{threeparttable}
\usepackage{booktabs}
\usepackage{adjustbox}
\usepackage{multicol}
\usepackage{lipsum}
\usepackage{graphicx}
\usepackage{booktabs}
\usepackage{caption}
\usepackage{pgfplots}
\usepackage{pgfplotstable}
\usepackage{subcaption}
\usepackage{tikz}
\usepackage{array}
\usepackage{pgfplots}
\newcolumntype{P}[1]{>{\centering\arraybackslash}p{#1}}

\title{\LARGE \bf
How Physics and Background Attributes Impact Video Transformers in Robotic Manipulation: A Case Study on Planar Pushing
}

\author{Shutong Jin, Ruiyu Wang, Muhammad Zahid and Florian T. Pokorny% <-this % stops a space
\thanks{The authors are with the School of Electrical Engineering and Computer Science, KTH Royal Institute of Technology
        {\tt\small shutong@kth.se; fpokorny@kth.se}. This work was partially supported by the Wallenberg AI, Autonomous Systems and Software Program (WASP) funded by the Knut and Alice Wallenberg Foundation. The computations were enabled by the supercomputing resource Berzelius provided by the National Supercomputer Centre at Linköping University and the Knut and Alice Wallenberg Foundation, Sweden.%
        }
}

\begin{document}

\maketitle
\thispagestyle{empty}
\pagestyle{empty}

%%%%%%%%%%%%%%%%%%%%%%%%%%%%%%%%%%%%%%%%%%%%%%%%%%%%%%%%%%%%%%%%%%%%%%%%%%%%%%%%
\begin{abstract}
    As model and dataset sizes continue to scale in robot learning, the need to understand how the composition and properties of a dataset affect model performance becomes increasingly urgent to ensure cost-effective data collection and model performance. In this work, we empirically investigate how physics attributes (color, friction coefficient, shape) and scene background characteristics, such as the complexity and dynamics of interactions with background objects, influence the performance of Video Transformers in predicting planar pushing trajectories. We investigate three primary questions: How do physics attributes and background scene characteristics influence model performance? What kind of changes in attributes are most detrimental to model generalization? What proportion of fine-tuning data is required to adapt models to novel scenarios? To facilitate this research, we present \textit{CloudGripper-Push-1K}, a large real-world vision-based robot pushing dataset comprising 1278 hours and 460,000 videos of planar pushing interactions with objects with different physics and background attributes. We also propose Video Occlusion Transformer (VOT), a generic modular video-transformer-based trajectory prediction framework which features 3 choices of 2D-spatial encoders as the subject of our case study. The dataset and source code are available at \href{https://cloudgripper.org}{https://cloudgripper.org}.
\end{abstract}

\section{Introduction}
The transformer architecture \cite{vaswani2017attention} has established itself as a highly effective tool across a diverse array of sequence-to-sequence problems in natural language processing \cite{vaswani2017attention,Radford2019LanguageMA,devlin-etal-2019-bert}, computer vision \cite{dosovitskiy2021an,liu2021swin,Arnab_2021_ICCV}, decision making \cite{NEURIPS2021_7f489f64,NEURIPS2022_b2cac94f,pmlr-v162-zheng22c} and robotic manipulation \cite{brohan2022rt, pmlr-v205-shridhar23a}. Video Transformers (VTs) \cite{liu2022video,NEURIPS2021_67f7fb87,arnab2021vivit} are widely studied for video-based downstream tasks and their ability to effectively capture complex spatio-temporal relationships and long-range dependencies makes them particularly intriguing for robot perception and planning, e.g., visually-grounded robotic manipulation through imitation learning \cite{pmlr-v155-dasari21a,10161288,wang2023mimicplay}, robotic video prediction \cite{rakhimov2020latent,gupta2023maskvit,nash2023transframer} and goal-conditioned planning with video demonstrations \cite{9712373,zhao2022p3iv,wang2023event}. However, VTs are computationally expensive \cite{khan2022transformers} and their training requires large video datasets \cite{tong2022videomae} that are usually costly to collect on real robots. Both the computational cost of training VTs from scratch and the requirement for large-scale data collection are challenges that may
 currently still hinder their widespread adoption in real-world robotic manipulation tasks. To alleviate this burden, a new set of empirical evidence is required to understand how the composition and properties of a dataset affect model performance and how much data is typically required for training and fine-tuning in the context of robotic manipulation. The impact of physics attributes (color, friction coefficient, shape) and scene background characteristics present in video datasets are aspects that are at present poorly understood in the context of emerging VT architectures for robotic manipulation. Since a rigorous theoretical framework describing the impact of these factors is at present unavailable to the robotics research community, our work proposes a first large-scale empirical case based on real-world data as a step towards developing an understanding on how these factors influence generic VT model performance in practice. 
  
%===================================================================
% FIG. 01
%===================================================================
 
\begin{figure}[!t]
    \centering
    \includegraphics[width=1\linewidth]{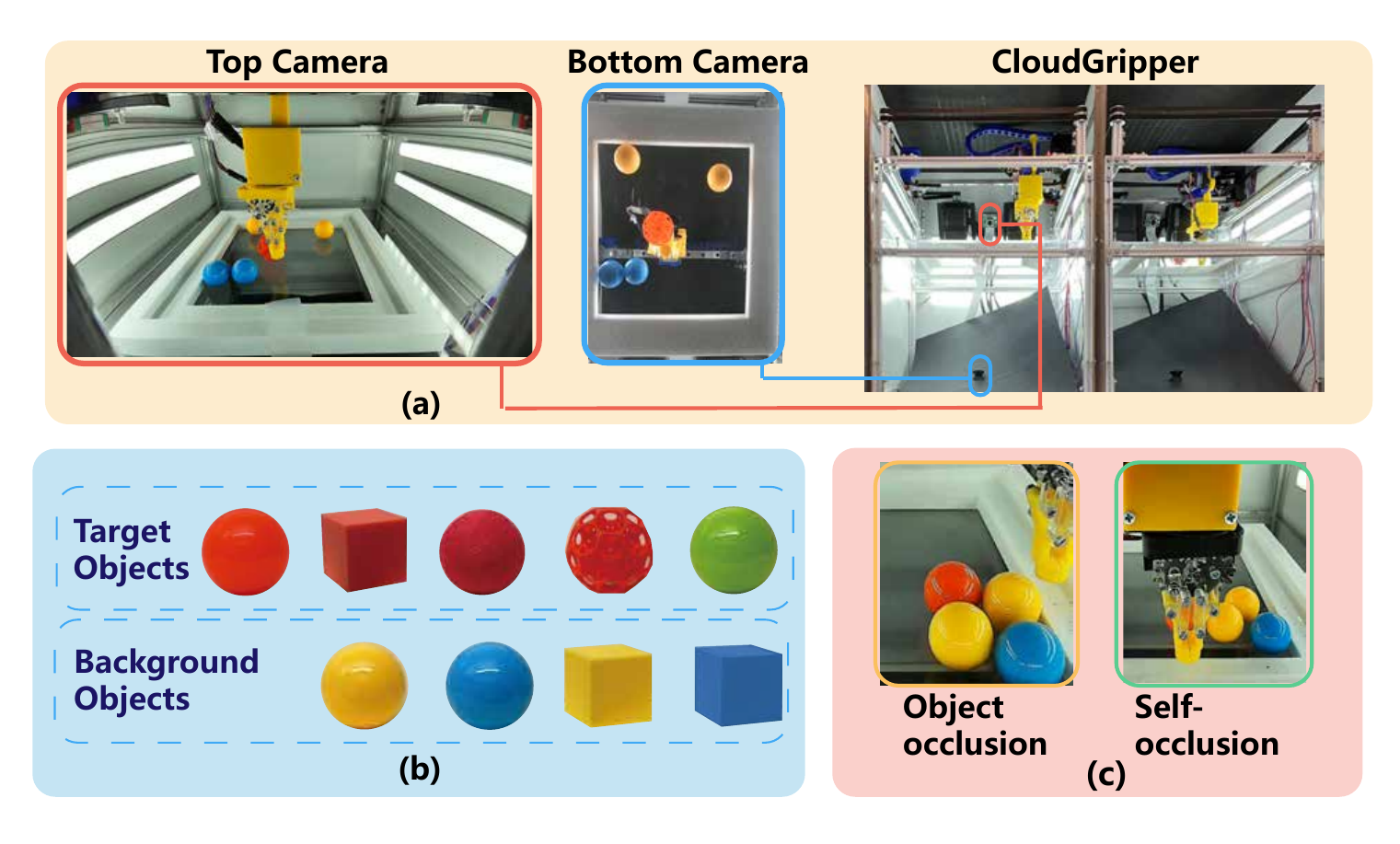}
    % \vspace{-2em}
    % sjin: maybe we should cite cloud gripper hardware paper?
    \caption{\small 
    An illustration of the data collection process on the \textit{CloudGripper} platform - an open source cloud robotics testbed with 32 robot arm cells. We collect planar robot pushing interaction videos from two camera views. (a) Collected top and bottom camera views and \textit{CloudGripper} robot cell side-view. (b) An illustration of target and background objects used in the case-study. (c) Illustration of object and gripper self-occlusion present in collected top camera videos.}
    \label{fig:dataset_introduction}
    \vspace{-1em}
\end{figure}
%=================================================================
Towards this end, we introduce \textit{CloudGripper-Push-1K}, a large real-world visual-based robot pushing dataset encompassing various target objects (\textit{Ball}, \textit{Cube}, \textit{Foam}, \textit{Icosahedron}) with distinct physics attributes (color, friction coefficient, shape). We furthermore vary the number and type of background objects that are placed in the same workspace as the primary object which the robot is interacting with. This approach creates variations in background complexity and physical dynamics due to secondary collisions during pushing interactions, resulting in 18 sub-datasets. The dataset is collected with the \textit{CloudGripper} system \cite{CloudGripper} and includes 1.4 TB, 1278 hours and 460,000 trajectories of robot pushing interactions, resulting in one of the largest robot datasets publicly available to date. We collect videos from a dual-camera system: a top camera view providing a scene overview with occlusions, and a bottom camera perspective that allows for an occlusion-free observation of object position as illustrated in \textit{Fig.}\ref{fig:dataset_introduction}(a). This video dataset, with its controlled variations in physics and background attributes will be made available and serves as a primary dataset resource for the presented case study. We believe that this dataset may also be useful for the community in understanding general occlusion problems and occlusion-aware networks with single camera input, providing an additional perspective distinct from other approaches assuming calibrated RGB-D \cite{shridhar2023perceiver} or multi-camera setups \cite{shridhar2022cliport}.
% =================================================================
% FIG. 02
% =================================================================
\begin{figure*}[ht!]
\centering
\includegraphics[width=1.0\linewidth]{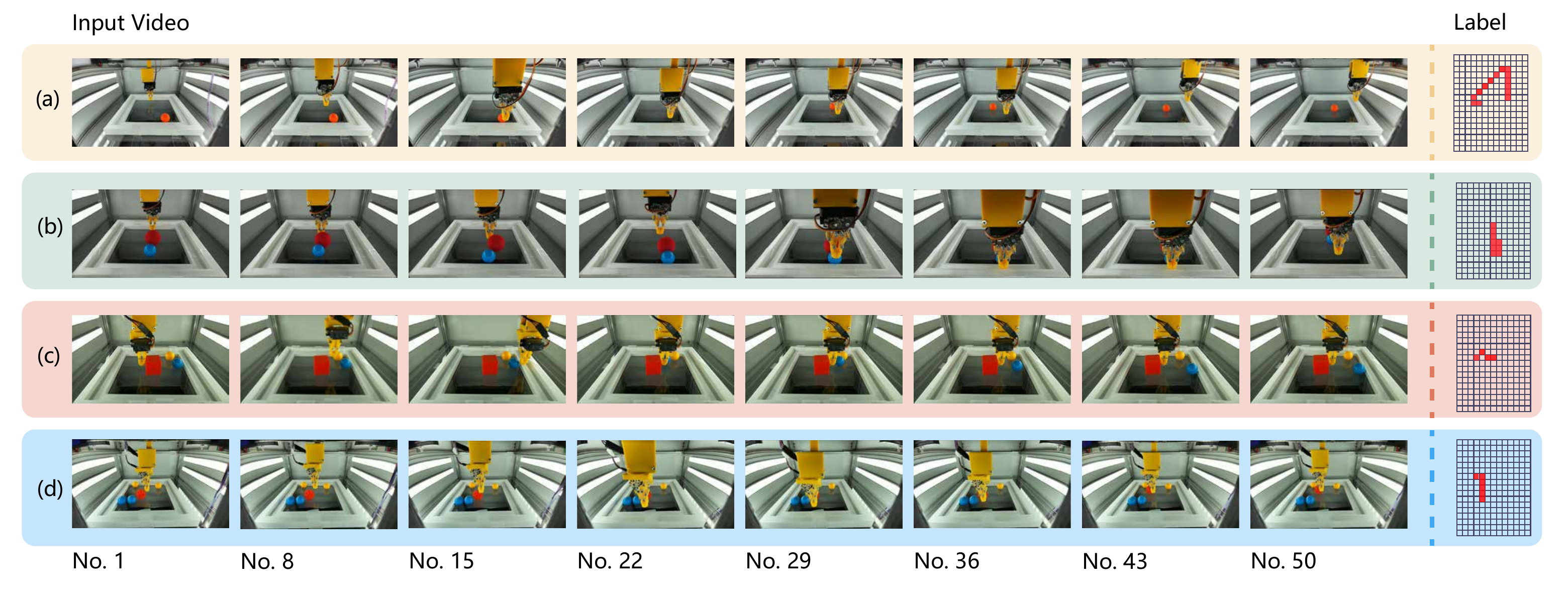}
\vspace{-1.5em}
\caption{\small An illustration of \textit{CloudGripper-Push-1K} with selected trajectories. (a) Target object \textit{Ball}. (b) Target object \textit{Foam} and one background object. (c) Target object \textit{Cube} and two background objects. (d) Target object \textit{Icosahedron} and four background objects.}
\label{fig:2_dataset}
\vspace{-1em}
\end{figure*}
% ================================================================
% \vspace{-0.1em}
% need polish

% Most of the Video transformers adopt the design of separate vision and temporal encoding. Therefore, we also present Video Occlusion Transformer (\textit{VOT}), a VT-based framework with a generic structure, which takes in the occluded top camera videos and predicts the trajectories of target objects in the coordinate of bottom camera view. 
In alignment with the generic approach outlined in VTN \cite{neimark2021video}, we designed Video Occlusion Transformer (\textit{VOT}), a modular framework consisting of a flexible 2D-spatial encoder (\textit{VOT-MaxViT}, \textit{VOT-MaxViT-2}, \textit{VOT-Swin-T}) and a temporal encoder. \textit{VOT} takes the occluded top camera videos as inputs and predicts the trajectories of target objects in the pixel coordinate of the bottom camera. To study the generalisability of VTs, the framework is trained from scratch across all tasks available in \textit{CloudGripper-Push-1K}. Zero-shot and few-shot learning are conducted for tasks with various physics attributes and scene background characteristics. The main contributions of this paper are: 
\begin{itemize}
    \item We present a large real-world video dataset \textit{CloudGripper-Push-1K} of 1278 hours of planar robot-object pushing interactions which enables the study of VTs and other vision-based large models for robotic manipulation. The dataset will be released as an open dataset to the community.   
    \item A generic Video Transformer framework \textit{VOT} covering three types of 2D-spatial encoders is designed for this specific robot-pushing setting to predict trajectories of target objects.
    \item We present ablation studies by training across 18 sub-datasets from scratch and performing zero-shot evaluation and fine-tuning on sub-datasets, providing empirical evidence on how physics attributes and scene background characteristics influence the performance of VTs.
\end{itemize}

% %%%%%%%%%%%%%%%%%%%%%%%%%%%%%%%%%%%%%%%%%%%%%%%%%%%%%%%%%%%%%%%%%%%%%%%%%%%%%%%%
% % =================================================================
% % FIG. 03
% % =================================================================

% \begin{figure*}[!t]
% \centering
% \includegraphics[width=0.9\linewidth]{figures/3_structure.pdf}
% \vspace{-0.5em}
% \caption{\small (a) Model Structure: Following the generic approach detailed in VTN\cite{neimark2021video}, our framework employs a modular design, starting with a 2D-spatial and followed by a temporal encoder. (b)Three types of 2D-spatial encoders adopted in the design. (c) Temporal encoder. (d)Three types of attention mechanisms used in the 2D-spatial encoders.}

% \label{fig:2_structure}
% \vspace{-1em}
% \end{figure*}
% % =================================================================

\section{Related Work}
\subsection{Model and Dataset Scalability}
% Scaling Large Models and Datasets, Large Model and Data Scaling
While transformers have shown significant potential across various application areas \cite{devlin-etal-2019-bert, liu2021swin, pmlr-v205-shridhar23a}, their quadratic complexity with respect to the number of available tokens imposes substantial challenges on both model training and data collection \cite{brohan2022rt}. Recent efforts have been initiated to investigate the scaling behavior with respect to model size, model depth, dataset scale and computing power \cite{tay2021scale, fedus2022switch, tay2022scaling, kaplan2020scaling}. These studies however mostly focused on the impact of the number of tokens on scaling and have not addressed how the composition and properties relating to the training dataset influence the generalisability of the models.

Some endeavors have been made to determine what information is most effective to be included in training data for model pre-training. \textit{Jing et al.} \cite{jing2023exploring} investigate the effects of pre-training datasets, model architectures and training methods on visual pre-training for robot manipulation. In \textit{KitchenShift} \cite{xing2021kitchenshift}, the authors study the zero-shot generalization of an imitation-based policy under domain shifts, such as lighting and camera pose. However, most of the aforementioned experiments and benchmarks \cite{xie2023decomposing, cobbe2020leveraging} are designed in simulated environments. For robotic manipulation, even high-fidelity simulations used today however still differ significantly from real-world video datasets both in terms of visual appearance and physics - particularly when it comes to friction and physical collisions in the environment, thus motivating additional empirical study of these topics on real world video data.

\subsection{Existing Video Datasets}
Most existing video datasets to date are designed for pure computer vision tasks, such as,  action recognition \cite{kay2017kinetics}, object tracking \cite{huang2019got} and video segmentation \cite{pont20172017}. Prominent examples include Kinetics400 and Kinetics600 \cite{kay2017kinetics}, Moments in Time \cite{monfort2019moments}, Something-Something V2 (SSv2) \cite{goyal2017something}, AViD \cite{piergiovanni2020avid}, Epic Kitchen \cite{Damen2022RESCALING} and Charades \cite{sigurdsson2018charades}. However, these datasets are not specifically tailored towards robotic manipulation settings. 

Recently, interest in video datasets for robotic manipulation has increased significantly, with initial datasets at larger scales such as RT-1 \cite{brohan2022rt}, HOPE-Video \cite{tyree2022hope} and BC-Z \cite{jang2022bc} being released. Nonetheless, owing to the traditionally high-cost associated with collecting large-scale real-world datasets, video-based robotic manipulation datasets today still largely remain either limited in size or only available to corporate actors. Moreover, a considerable number of datasets are gathered in simulated environments \cite{robomimic2021, fu2020d4rl, wong2022error, james2019rlbench, mandlekar2018roboturk}, thereby suffering from the sim-to-real gap \cite{hofer2021sim2real}. 

The dataset \textit{CloudGripper-Push-1K} presented in this work therefore fills an important gap by providing large real-world video data of planar pushing interactions comprising 460,000 trajectories across a number of settings and serves as a foundational dataset that allows us to control for changes in a single variable to conduct experimental evaluations.

\subsection{Video Transformers}
The transformer architecture, originally designed for natural language processing, has shown a strong capacity when being adapted to vision tasks \cite{chen2021crossvit, 9711179}. Video transformers (VTs) extend the structure of vision transformers \cite{dosovitskiy2020image} by incorporating the temporal dynamics inherent to video data and have achieved impressive results on video classification \cite{arnab2021vivit, liu2022video, neimark2021video}, robotic manipulation \cite{brohan2022rt, shridhar2023perceiver, brohan2023rt} and object tracking tasks \cite{xie2023videotrack, lin2022swintrack, wang2021transformer}.

To address the challenges of computational intensity and to capture the rich information introduced by the temporal dimension, various factorization methods have been adopted in recent VTs. Notably, Swin Video Transformer \cite{liu2022video} adapts the Swin Transformer \cite{liu2021swin} to include inductive bias, hierarchical structures, and translational invariance, thereby integrating the advantages of Convolutional Neural Networks into transformers. ViViT \cite{arnab2021vivit} presents several strategies for factorizing both spatial and temporal dimensions to manage the long token sequences in videos. VTN \cite{neimark2021video} utilizes a 2D-spatial backbone for feature extraction followed by a temporal encoder \cite{beltagy2020longformer} to realize flexible video processing. In our work, the proposed \textit{VOT} framework adopts the generic structure proposed by VTN.

\section{Dataset}

\begin{table}[!t]
    \centering
    \caption{Dataset Overview}
    \begin{threeparttable}
    \begin{tabular}{cccc}
         \toprule
         \textbf{Object} & \textbf{Background} & \textbf{Training} & \textbf{Testing} \\
         \midrule
         \textbf{Ball (Red)} & \text{Single} & 30030 & 2800 \\
          & \text{Double} & 22860 & 1881\\
          & \text{Triple} & 20720 & 2570\\
          & \text{Quintuple} & 20000 & 1696\\
         \midrule
         \textbf{Cube} & \text{Single} & 21860 & 1889\\
          & \text{Double} & 32870 & 1500\\
          & \text{Triple} & 27800 & 2311\\
          & \text{Quintuple} & 21610 & 2180\\
         \midrule
          \textbf{Foam} & \text{Single} & 19450 & 2163\\
          & \text{Double} & 20440 & 2134\\
          & \text{Triple} & 20838 & 2100\\
          & \text{Quintuple} & 21430 & 2129\\
         \midrule
          \textbf{Icosahedron} & \text{Single} & 21670 & 2697\\
          & \text{Double} & 24980 & 2379\\
          & \text{Triple} & 21430 & 3150\\
          & \text{Quintuple} & 22070 & 1661\\
          & \text{Quintuple\_Static} & 25360 & 3392\\
         \midrule
          \textbf{Ball (Green)} & \text{Single} & 21540 & 1660\\
          \bottomrule
    \end{tabular}
    \begin{tablenotes}    
    \footnotesize 
        \item This table outlines the sub-tasks and number of videos in each of the 18 training and testing sub-datasets. The background column indicates the total number of objects present in the given setting.
    \end{tablenotes}
    \end{threeparttable}
    \label{tab:dataset/stats}
    \vspace{-1em}
\end{table}

%%%%%%%%%%%%%%%%%%%%%%%%%%%%%%%%%%%%%%%%%%%%%%%%%%%%%%%%%%%%%%%%%%%%%%%%%%%%%%%%
% =================================================================
% FIG. 03
% =================================================================
\begin{figure*}[ht]
\centering
\includegraphics[width=0.8\linewidth]{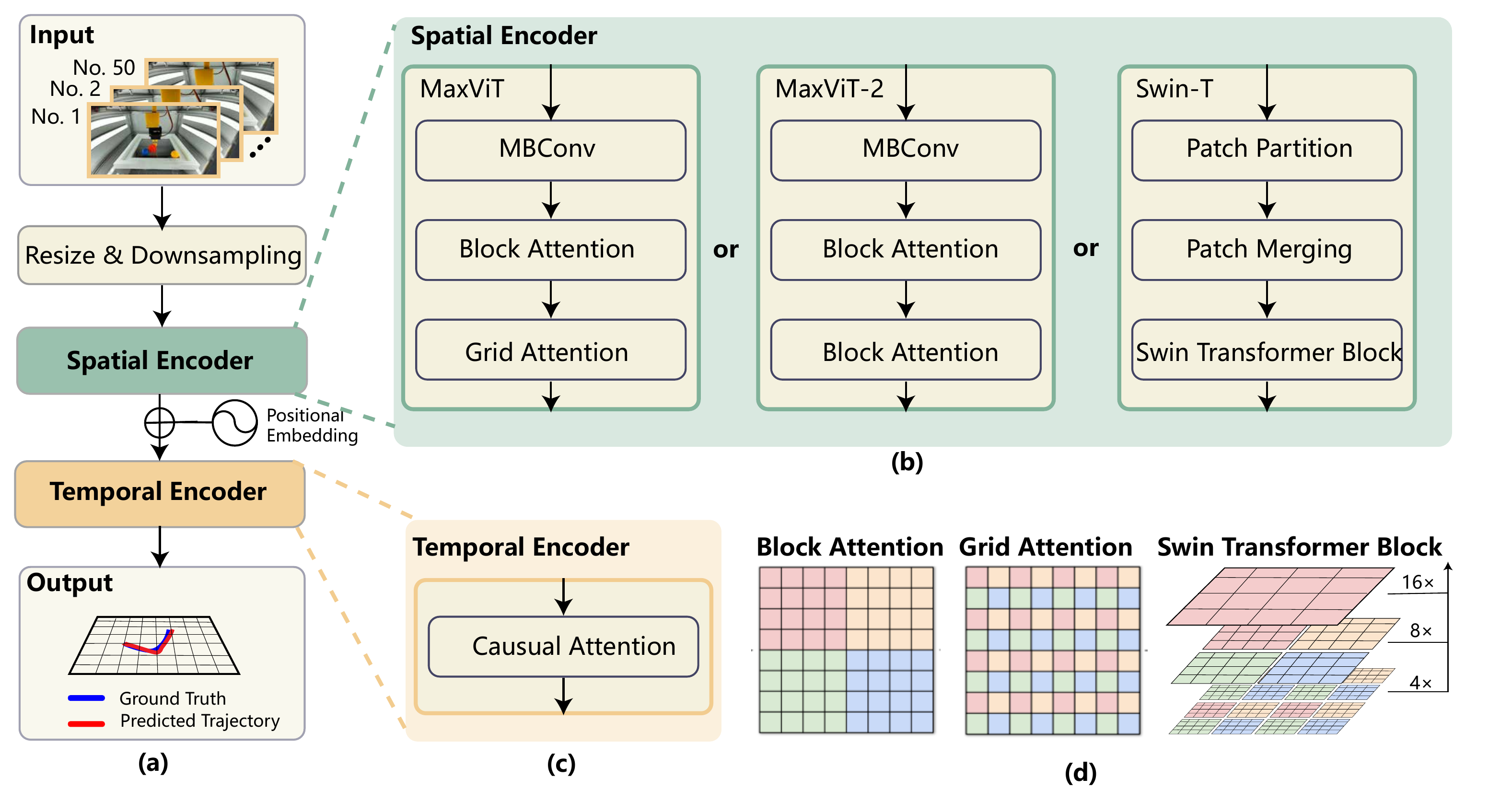}
\vspace{-0.5em}
\caption{\small (a) VOT Model Structure: Following the generic approach detailed in VTN \cite{neimark2021video}, our framework employs a modular design, starting with a 2D-spatial and followed by a temporal encoder. (b) The three types of 2D-spatial encoders that were adopted in the design. (c) Temporal encoder, (d) The three types of attention mechanisms used in the 2D-spatial encoders.}

\label{fig:2_structure}
\vspace{-1em}
\end{figure*}
% =================================================================

The \textit{CloudGripper-Push-1K} dataset is collected using the open cloud-robotics platform, \textit{CloudGripper} \cite{CloudGripper} developed at KTH. The dataset provides a collection of 1.4 TB of data with 1278 hours and 460,000 trajectories of robot pushing RGB videos varying the target objects from ball, cube, icosahedron (a convex polyhedron with 20 faces) to a foam ball, and with background scene settings with a varying number of rigid balls or cubes, as shown in \textit{Fig.}\ref{fig:2_dataset}.  In the dataset, each trajectory $\textit{Traj} = (V_{t},\; V_{b},\; P_{o})$ consists of:
\begin{enumerate}
\item A high-resolution occluded top-camera video $V_{t}$ of dimension $T \times H_{t} \times W_{t} \times C$.
\item A corresponding occlusion-free bottom-camera video $V_{b}$ of dimension $T \times H_{b} \times W_{b} \times C$.
\item A ground-truth trajectory for the target object in pixel space, $P_{o}\in\{ (i,\;x,\;y)\mid i,\;x,\;y\in\mathbb{Z},\;0\leq i \leq T, \;0\leq x \leq H_{b},\; 0\leq y \leq W_{b}\}$.
\end{enumerate}
where $H_{t}$, $W_{t}$, $H_{b}$, $W_{b}$ are equal to 1280, 720, 480 and 640 respectively, indicating the heights and widths of $V_{t}$ and $V_{b}$ frames; $T=50$ is the number of frames collected for each video and $C=3$, representing the number of video color channels.
The dataset is collected by moving the robot gripper, with a moving range of $15 \;cm \times 15\;cm$, and executing random planer pushing of target objects in the workspace environments. During the pushing process, two RGB videos are recorded with our dual-camera system at a frame rate of 5 fps and saved when the total number of frames reaches 50 in a video. The bottom camera view is observed through a transparent plexiglass plate from below the planar pushing surface. The ground-truth trajectories $P_{o}$ are later derived from $V_{b}$ using a RGB color-based tracking algorithm recording the pushed object's motions. 

We design 18 sub-datasets with varying selections of target objects with different physics attributes and various scene backgrounds. An overview of target and background objects can be seen in \textit{Fig.}\ref{fig:dataset_introduction}(b). Four target objects (\textit{Ball (red)}, \textit{Cube}, \textit{Foam}, \textit{Icosahedron}) being pushed in workspace backgrounds with zero (\textit{Single}), one (\textit{Double}), two (\textit{Triple}) and four background objects (\textit{Quintuple}) are captured, resulting in 16 sub-datasets. The red ball (made from plastic) and foam object (also a ball) share identical shape and color but vary in friction coefficient. The cube and icosahedron are 3D printed from identical plastic material and only vary in shape. These two groups of sub-datasets are designed to investigate the impact of friction coefficient and shape. Additionally, another two sub-datasets, pushing a single green \textit{Ball} without any other objects in the scene and pushing an icosahedron with 4 cubes in the background (\textit{Quintuple\_Static}), have been created to examine the impact of target object color (compared with \textit{Single Ball}) and background dynamics (compared with \textit{Icosahedron Quintuple}, since the background objects are balls whose rolling movements after the collision are more dynamic than cubes). Each sub-dataset consists of approximately 20,000 training and 2,000 test trajectories. \textit{Table.}\ref{tab:dataset/stats} outlines the sub-task settings and number of videos in each training and testing sub-dataset.

Finally, we would like to emphasize that the presence of frequent occlusions both due to occlusion behind the gripper and due to occlusions by other objects which presents an interesting challenge to training a model based on top camera videos in \textit{CloudGripper-Push-1K}, examples of such occlusions are visible in \textit{Fig.}\ref{fig:dataset_introduction}(c).

\section{Model}

The overall design of \textit{VOT} is illustrated in \textit{Fig.}\ref{fig:2_structure}. Our framework is modular in construction which aligns with the generic approach outlined in VTN \cite{neimark2021video}, incorporating a flexible 2D-spatial encoder for frame feature extraction, followed by a temporal encoder to enhance inference from previous frames. In \textit{VOT}, we adopt three vision transformers as the 2D-spatial encoders, enabling a broader investigation into the generalisability of video transformers.

\textbf{Input.} The input to \textit{VOT} is a top camera video $V_{t}$ with 50 frames. We preprocess the inputs by downsampling them to 25 frames and resizing the frames to \( 224 \times 224 \) pixels.

\textbf{Spatial Encoder.} In \textit{VOT}, three vision transformer architectures are adopted as the 2D-spatial encoder. MaxViT \cite{tu2022maxvit}, enhances the original ViT model \cite{dosovitskiy2021an} with reduced computational complexity by incorporating Block Attention and Grid Attention (\textit{VOT-MaxViT}). Block Attention is designed to capture local features, while Grid Attention targets global information. 

Questioning the necessity of global attention and whether a stronger focus on local features could enhance performance in the robotic setting (where more focus on the interaction between gripper and object is desired), we design \textit{VOT-MaxViT-2}, a variant of \textit{VOT-MaxViT}, by changing the second attention in the original MaxViT from Grid Attention to Block Attention. Additionally, we also adopt Swin Tranformer \cite{liu2021swin}, a well-established architecture with hierarchical structure and shifted window for self-attention, as another variant of the spatial encoder (\textit{VOT-Swin-T}). By incorporating three self-attention mechanisms into the 2D-spatial encoder backbone, we aim to investigate the impact of physics and background attributes on these model design choices in VTs.
% % =================================================================
% % FIG. 03
% % =================================================================
% \begin{figure*}[!ht]
% \centering
% \includegraphics[width=0.8\linewidth]{figures/3_structure.pdf}
% \vspace{-0.5em}
% \caption{\small (a) VOT Model Structure: Following the generic approach detailed in VTN \cite{neimark2021video}, our framework employs a modular design, starting with a 2D-spatial and followed by a temporal encoder. (b) Three types of 2D-spatial encoders adopted in the design. (c) Temporal encoder. (d) Three types of attention mechanisms used in the 2D-spatial encoders.}

% \label{fig:2_structure}
% \vspace{-1em}
% \end{figure*}
% % =================================================================
\textbf{Temporal Encoder.} The temporal encoder follows the initial self-attention mechanism proposed by \textit{Vasawani et al.} \cite{vaswani2017attention} to learn the temporal relationships across frames, with causal attention masking out future information.

\textbf{Output.} The output of \textit{VOT}, with a dimension of $12 \times 16$, represents the predicted target object trajectory traversed within the 25 frames in a down-sampled pixel coordinate of the bottom camera. We adopt the down-sampling considering the trade-off between computational limitation and necessary prediction resolution (small object moving range of $12\times 12 cm$).

\input{tables}

\section{Experiments}
\subsection{Implementation Details}
\textbf{Training.} We train and evaluate 3 variants of \textit{VOT} (\textit{VOT-MaxViT}, \textit{VOT-MaxViT-2}, and \textit{VOT-Swin-T}) on each of the 18 \textit{CloudGripper-Push-1K} sub-datasets from scratch, where the sizes of training and testing sets are 20,000 and 2,000 respectively. The models are diversified by object type (\textit{Ball} (Red), \textit{Cube}, \textit{Foam}, \textit{Icosahedron}) and background complexity (\textit{Single}, \textit{Double}, \textit{Triple}, \textit{Quintuple}) in training data. In total, 48 models are formed by combining these factors, along with 6 additional models, trained on \textit{Ball\_Green} and \textit{Quintuple\_Static} to explore other attributes (color, background dynamics). Training is conducted with 8 NVIDIA A100 GPUs on the Berzelius supercomputer \cite{Berzelius}. Each model is trained for 100 epochs with a batch size equals to 4, requiring approximately 12 hours of training. We utilize the Adam optimizer \cite{kingma2014adam} with constant learning rate of $1 \times 10^{-5}$ for \textit{VOT-Swin-T} and $1 \times 10^{-4}$ for \textit{VOT-MaxViT} and \textit{VOT-MaxViT-2}. All models utilize random initialization and the Mean Squared Error (MSE) is adopted as loss function. The \textit{VOT-MaxViT}, \textit{VOT-MaxViT-2}, \textit{VOT-Swin-T} comprise $5.2 \times 10^{7}$, $5.2 \times 10^{7}$, $4.9 \times 10^{7}$ parameters respectively. % need modifications

\textbf{Testing.} Each of the 18 sub-datasets includes a separate testing set consisting of around 2,000 videos. All zero-shot experiments are conducted on the testing set, where the same robot is used as during the collection of the original training data. This ensures that performance metrics are not influenced by external disturbances and minor differences between individual \textit{CloudGripper} robot instances as these are unrelated to the attributes under study.

\textbf{Evaluation Metrics.} Two metrics are employed to assess the performance and generalisability of \textit{VOT}s. The Prediction Error ($PE$) quantifies the discrepancy between the predicted and ground-truth trajectories. The Generation Gap ($GP$) evaluates the model's generalisability when applied to datasets with different attributes by taking the difference between the model's current $PE_{new}$ and the sub-dataset's $PE_{previous}$ of training from scratch.
\begin{equation}
    Prediction\;Error\;(PE) = \frac{1}{M\times N}\sum_{i=1}^{M}\sum_{j=1}^{N} (R_{ij} - P_{ij})^2
\end{equation}
\begin{equation}
    Generation\;Gap\;(GP) = PE_{new} - PE_{previous},
\end{equation}
where $M=12$ and $N=16$ are the height and width of the network outputs; \( P \) represents the predicted trajectory; \( R \) denotes the downsampled ground-truth trajectory.
\subsection{Experiments with Varying Scene Background Characteristics}
\textbf{Complexity.} As a proxy for the complexity of real-world scene backgrounds, we vary the number of objects placed in the workspace. In this context, ``background object'' refers to items presented in the scene that are not the target object being pushed by the robot gripper. We experiment with 16 sub-datasets, analyzing 4 levels of background complexity in combination with 4 distinct target objects being pushed: \textit{Ball}, \textit{Cube}, \textit{Foam} and \textit{Icosahedron}. \textit{VOT}s are trained from scratch and the zero-shot generalisability of models trained on a dataset without background objects are then tested on all other environments. 
The \textit{PE}s in \textit{Table.}\ref{tab1} are higher for the \textit{Double} and \textit{Triple} settings compared with a \textit{Single} background object for most cases (21 out of 24), which shows that complex backgrounds cause difficulties for training of VTs and the model performance drops even when performing the same pushing tasks. However, all of the 3 variants behave well in \textit{Quintuple} data. We hypothesize this is a consequence of the constrained moving area for the target object with the introduction of an increasing number of background objects. The non-zero \textit{GP}s in \textit{Table.}\ref{tab1} show that i) VT's performance is sensitive to the presence of background objects; ii) in \textit{Ball} and \textit{Icosahedron}, the \textit{GP}s increase with background complexity, showing a decline in generalisability with complex backgrounds. In the case of \textit{Cube} and \textit{Foam}, this trend is relatively weaker - possibly due to the reduced amount of dynamic interaction in the dataset (as the Cube and Foam objects do not roll and come to rest quickly). 

\textbf{Dynamics.} In an additional experiment, \textit{Quintuple\_Static}, we replace balls in the background with cubes of the same size and color in \textit{Icosahedron}'s \textit{Quintuple} setting. Since the cubes slide rather than roll and move less dramatically compared with balls,
this experiment serves to assess the impact of environmental dynamics on the VT's performance. The last two columns of \textit{Table.}\ref{tab2} confirm the intuition that the dynamic environments are harder to predict and VT models trained on these datasets tend to have a better zero-shot generalisability on settings with more static backgrounds. 

Finally, in \textit{Table.}\ref{tab1}, it can be observed that \textit{VOT-MaxViT} generally outperforms \textit{VOT-MaxViT-2} (12 out of 17 cases). This observation underscores the importance of retrieving more global information through Grid Attention, even in robotic manipulation settings where the interaction between grippers and objects is the primary focus.

\subsection{Experiments with Varying Physics Attributes}
Here we present a zero-shot generalisability study of physics attributes and scene background characteristics as summarized in \textit{Table.}\ref{tab2}.

\textbf{Color.} We compare the zero-shot performance of models trained on data with a single green \textit{Ball} or red \textit{Ball}. The \textit{GP} values for this group are the highest among the four controlled experiments, illustrating that VTs are highly sensitive to object color. This can be explained by the fact that shifts in object color dramatically affect pixel value of the input videos thus causing a significant perturbation in the inputs of the VOT models. Pure appearance factors as simple as color which should not affect trajectory prediction in the robotic manipulation context hence can still cause significant adverse effects on VT's performance.

\textbf{Friction.}
As mentioned before, \textit{Foam} and \textit{Ball} share near identical shape and color but \textit{Ball} features a lower friction coefficient compared with \textit{Foam}. Therefore, we compare the performance of this group to understand the impact of friction. When the force on the gripper is large enough to overcome the static friction, the observed effect of the differing friction coefficients is that objects with lower rolling friction coefficient, such as \textit{Ball}, start rolling on the plane and are bounced back from the workspace boundary, resulting in longer and more unpredictable trajectories. We observe asymmetries on the \textit{GP}s of \textit{VOT-MaxViT} and \textit{VOT-MaxViT-2}, which reveal that it may be easier for models trained on longer trajectories to generalize. However, this conclusion does not hold true for \textit{VOT-Swin-T}. This could potentially result from the different architectures between MaxViT \cite{tu2022maxvit} and Swin Transformer \cite{liu2021swin}. The Swin Transformer in particular has a more local focus with strong inductive bias, complicating the prediction of a fast-moving object (\textit{Ball}).

\textbf{Shape.} Here we consider the effect of changing object shapes with a comparative study of \textit{Cube} and \textit{Icosahedron}, which are both 3D printed from the same plastic material. Based on the average \textit{GP}s across three controlled physics attribute groups, it appears that color has a greater impact than shape on performance, while shape and friction show a more comparable level of influence on \textit{GP}s.

% \paragraph*{Experiment of Saturation Curve}
% As shown in \textit{Fig.}\ref{fig4:saturation_curve}, we train models for all objects with dataset size of 500, 1000, 10000, 15000, 20000 videos respectively. ``Elbow points'' in $PE$ are identified across four objects, indicating that the performance of transformer architectures may saturate with sufficient data on single tasks. This provides valuable guidance for optimizing resource allocation, reducing the computational overhead of training transformers.

\subsection{Fine-Tuning and Qualitative Observations}
% \input{plot_finetune}
%===================================================================
% FIG. 05
%===================================================================
 
\begin{figure}[!t]
    \centering
    \includegraphics[width=0.95\linewidth]{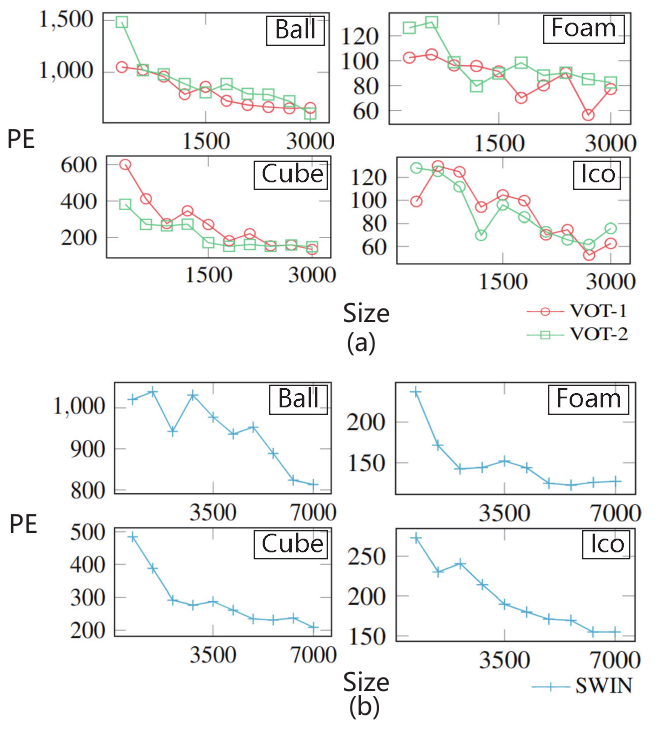}
    % \vspace{-2em}
    % sjin: maybe we should cite cloud gripper hardware paper?
    \caption{\small \textit{PE} vs fine-tuning dataset size graphs. Models are pre-trained for 90 epochs on \textit{Single} dataset and then fine-tuned on \textit{Quintuple} data for 10 epochs and evaluated on \textit{Quintuple}. The x-axis represents the number of the fine-tuning videos. (a) Fine-tuning curves of \textit{VOT-MaxViT} (VOT-1) and \textit{VOT-MaxViT-2} (VOT-2) with x-axis ranging from 300 to 3000. (b) Fine-tuning curves of \textit{VOT-Swin-T} (SWIN) with x-axis ranging from 700 to 7000.}
    \label{fig5:fine_tune_curve}
    \vspace{-1em}
\end{figure}
%=================================================================

Since the zero-shot generalisability among the presented settings is not satisfying for many cases, we would like to investigate whether it is possible to reach the performance level of a model that is trained from scratch by means of fine-tuning with a limited set of data. 

We therefore fine-tune pre-trained models for 10 epochs with varying sizes of additional data, ranging from 300 to 3000 videos for \textit{VOT-MaxViT} and \textit{VOT-MaxViT-2} and 700 to 7000 videos for \textit{VOT-Swin-T}, as shown in \textit{Fig.}\ref{fig5:fine_tune_curve}. While it is commonly accepted that smaller fine-tuning datasets are preferred to leverage pre-trained models effectively, our results indicate that although \textit{PE} generally decreases with larger dataset sizes, when the fine-tuning dataset remains relatively small, the model performance might even decline with the increasing dataset size. Moreover, achieving comparable performance to MaxViT \cite{tu2022maxvit} in this scenario demands a substantially larger fine-tuning dataset when using Swin Transformer \cite{liu2021swin} as 2D-spatial encoder. 

%===================================================================
% FIG. 01
%===================================================================
 
\begin{figure}[!t]
    \centering
    \includegraphics[width=0.7\linewidth]{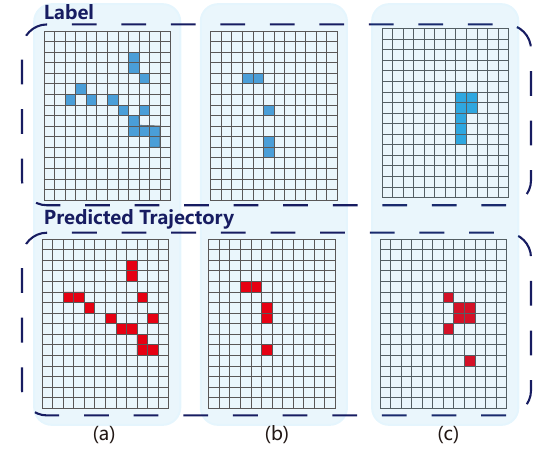}
    % \vspace{-2em}
    % sjin: maybe we should cite cloud gripper hardware paper?
    \caption{\small Examples of predicted and corresponding ground-truth trajectories, with \textit{Ball} being the target object. (a) A complicated long trajectory of an object colliding with the boundary four times. (b) A successful prediction that featured heavy occlusion in the top camera view. (c) A failure mode. Note that the discontinuities in labels and predicted trajectories are a result of the downsampling of input videos. Refer to the supplementary video for more details.}
    \label{fig:4_examples}
    \vspace{-1em}
\end{figure}
%=================================================================
In \textit{Fig.}\ref{fig:4_examples}, we present examples of predicted and ground truth trajectories - while VTs are able to predict complex long (a) or occluded (b) trajectories, failure cases also present in the case of (c). In (c), the target object (\textit{Ball}) 
exhibits an interesting unusual oscillation motion pattern possibly due to scratches on the plexiglass plate, representing a case not seen during training. Videos of examples described above and a data saturation curve of \textit{VOT-MaxViT} can be found in the supplementary video.

% Saturation curves in terms of number of training data across four objects containing ``Elbow points'' can be found in Fig.\ref{fig4:saturation_curve} and indicate that the dataset is of sufficient size on a per task basis.

% \input{plot_sat}

\subsection{Limitations and Conclusions}
While we have presented empirical results backed by a large real-world dataset, various factors such as the down-sampling of frame rate and resolution that was necessary for computational complexity reasons may affect our observations. Furthermore, we focus on typical VT architecture design components in the specific narrow setting of planar pushing in order to be able to report results at large data scales. Our experiments across the data subsets provide empirical evidence that: i) When training from scratch, VTs' performance appears to decline with increasing background scene complexity. Furthermore, models trained on dynamic backgrounds exhibit enhanced generalisability as compared to static backgrounds in zero-shot scenarios. ii) Among the physics attributes considered (color, friction coefficient, shape), the VTs' performance displays the greatest sensitivity to color - which is not a design goal in the context of robotic manipulation. Notably, shape and color demonstrate symmetry in cross zero-shot generalisability tests, while friction exhibits asymmetry, potentially due to the model architecture. iii) Regarding the fine-tuning process, we observed variations in optimal fine-tuning dataset sizes across settings and models. Furthermore, we observe that fine-tuning may even have a negative effect on model performance if the dataset size is not large enough.

% \addtolength{\textheight}{-12cm}   % This command serves to balance the column lengths
%                                   % on the last page of the document manually. It shortens
%                                   % the textheight of the last page by a suitable amount.
%                                   % This command does not take effect until the next page
%                                   % so it should come on the page before the last. Make
%                                   % sure that you do not shorten the textheight too much.

%%%%%%%%%%%%%%%%%%%%%%%%%%%%%%%%%%%%%%%%%%%%%%%%%%%%%%%%%%%%%%%%%%%%%%%%%%%%%%
% \newpage
\bibliographystyle{IEEEtran}
\balance
\bibliography{citations}
\end{document}

%% file: tables.tex
\begin{table*}[!t]
	\caption{Scratch and Zero-shot Performance of \textit{VOT} on \textit{CloudGripper-Push-1K} Dataset}
	\centering
	\label{tab1}
        % Adjusting column spacing
        \setlength{\tabcolsep}{1.5pt} % default is 6pt
        % Adjusting row spacing
        \renewcommand{\arraystretch}{1.3} % default is 1'
        \begin{threeparttable}
        \begin{tabular}{rc|cccc|cccc|cccc|ccccc}
            \toprule
            {} & {} & \multicolumn{4}{c|}{\textbf{Ball}} & \multicolumn{4}{c|}{\textbf{Cube}} & \multicolumn{4}{c|}{\textbf{Foam}} & \multicolumn{5}{c}{\textbf{Icosahedron}}\\
            {} & {} & {Sig.} & {Dou.} & {Tri.} & {Quin.} & {Sig.} & {Dou.} & {Tri.} & {Quin.} & {Sig.} & {Dou.} & {Tri.} & {Quin.} & {Sig.} & {Dou.} & {Tri.} & {Quin.} & {Quin\_S}\\
            \midrule
            \textbf{VOT-MaxViT}& {} & 413/{--} & 512/134 & 481/132 & 218/858 & 15/{--} & 65/191 & 33/730 & 28/639 & 62/{--} & 164/465 & 129/443 & 55/232 & 33/{--} & 42/225 & 25/368 & 23/591 & 30/104\\
            \textbf{VOT-MaxViT-2}& {} &  421/{--} & 542/18 & 455/1042 & 160/1404 & 20/{--} & 62/659 & 36/601 & 28/538 & 63/{--} & 174/345 & 127/303 & 56/180& 34/{--} & 46/127 & 24/224 & 24/272 & 27/367\\
            \textbf{VOT-Swin-T}& {} &  508/{--} & 592/414 & 685/887 & 300/1214 & 23/{--} & 80/183 & 40/806 & 36/785 & 97/{--} & 209/467 & 151/455 & 73/368& 55/{--} & 58/310 & 30/458 & 28/465 & 33/521\\
            \bottomrule
        \end{tabular}
    \begin{tablenotes}    
    \footnotesize 
        \item The results are shown in the format of $PE$/$GP$ indicating Prediction Error and Generation Gap. $GP$, as defined in \textit{Equation} 2, measures the zero-shot performance of models trained with the \textit{Single} dataset, for each target object, on those with a higher number of objects. \textit{GP}s are calculated by deducting \textit{PE} of the model trained from scratch from \textit{PE} of models trained on \textit{Single}. $Quin\_S$ stands for $Quintuple\_Static$.
    \end{tablenotes}
    \end{threeparttable}
\vspace{-0.1em}
\end{table*}

\begin{table*}[!t]
	\caption{Study of Physics Attributes and Background Characteristics}
	\centering
	\label{tab2}
        \setlength{\tabcolsep}{1.5pt} % default is 6pt
        \renewcommand{\arraystretch}{1.3} % default is 1'
        \begin{threeparttable}
        \begin{tabular}{rc|P{1.25cm}P{1.25cm}|P{1cm}P{1.5cm}|P{1.25cm}P{1.25cm}|P{1.25cm}P{1.25cm}}
            \toprule
            {} & {} & \multicolumn{2}{c|}{\textbf{Friction}} & \multicolumn{2}{c|}{\textbf{Shape}} & \multicolumn{2}{c|}{\textbf{Color}} & \multicolumn{2}{c}{\textbf{Background}}\\
            {} & {} & {Foam} & {Ball} & {Cube} & {Icosahedron} & {Ball\_R} & {Ball\_G} & {Dynamic} & {Static}\\
            \midrule
            \textbf{VOT-MaxViT}& {} & 1033 & 203 & 654 & 612 & 1098 & 1509 & 186 & 13\\
            \textbf{VOT-MaxViT-2}& {} &  1308 & 783 & 769 & 780 & 1372 & 1538 & 38 & 21\\
            \textbf{VOT-Swin-T}& {} &  554 & 1103 & 869 & 973 & 1485 & 1590 & 164 & 73\\
            \midrule
            \textbf{Average} & {} & 965 & 696 & 764 & 788 & 1318 & 1546 & 129 & 36\\
            \bottomrule
        \end{tabular}
    \begin{tablenotes}    
    \footnotesize 
        \item The table presents the values of zero-shot $GP$. $GP$ reflects the zero-shot generalisability of the other model in the same controlled group. For example, the $GP$ for \textit{Foam} in group \textit{Friction} equals to $PE$ of model trained on the \textit{Foam} dataset and tested on \textit{Ball} minus $PE$ of model trained on \textit{Ball} and tested on \textit{Ball}. $Ball\_R$ and $Ball\_G$ stand for red ball and green ball respectively.
    \end{tablenotes}
    \end{threeparttable}
\vspace{-1em}
\end{table*}